\documentclass{article}

\usepackage[nonatbib,final]{nips_2017}

\usepackage[utf8]{inputenc} 
\usepackage[T1]{fontenc}    
\usepackage{hyperref}       
\usepackage{url}            
\usepackage{booktabs}       
\usepackage{amsfonts}       
\usepackage{nicefrac}       
\usepackage{microtype}      
\usepackage{cite}
\usepackage{amsmath}
\usepackage{graphicx}
\newcommand{\cev}[1]{\reflectbox{\ensuremath{\vec{\reflectbox{\ensuremath{#1}}}}}}

\title{Improving Visually Grounded Sentence Representations with Self-Attention}

\author{ 
  Kang Min Yoo, Youhyun Shin, Sang-goo Lee\\
  Deparment of Computer Science\\
  Seoul National University\\
  \texttt{\{kangminyoo, shinu89, sglee\}@europa.snu.ac.kr} \\
}

\begin{document}

\maketitle

\begin{abstract}
Sentence representation models trained only on language could potentially suffer from the grounding problem. Recent work has shown promising results in improving the qualities of sentence representations by jointly training them with associated image features. However, the grounding capability is limited due to distant connection between input sentences and image features by the design of the architecture. In order to further close the gap, we propose applying self-attention mechanism to the sentence encoder to deepen the grounding effect. Our results on transfer tasks show that self-attentive encoders are better for visual grounding, as they exploit specific words with strong visual associations.
\end{abstract}

\section{Introduction}

Recent NLP studies have thrived on distributional hypothesis. More recently, there have been efforts in applying the intuition to larger semantic units, such as sentences, or documents. However, approaches based on distributional semantics are limited by the \textit{grounding problem} \cite{harnad1990symbol}, which calls for techniques to ground certain conceptual knowledge in perceptual information.

Both NLP and vision communities have proposed various multi-modal learning methods to bridge the gap between language and vision. However, how general sentence representations can be benefited from visual grounding has not been fully explored yet. Very recently, \cite{kiela2017learning} proposed a multi-modal encoder-decoder framework that, given an image caption, jointly predicts another caption and the features of associated image. The work showed promising results for further improving general sentence representations by grounding them visually. However, according to the model, visual association only occurs at the final hidden state of the encoder, potentially limiting the effect of visual grounding.

Attention mechanism helps neural networks to focus on specific input features relevant to output. In the case of visually grounded multi-modal framework, applying such attention mechanism could help the encoder to identify visually significant words or phrases. We hypothesize that a language-attentive multi-modal framework has an intuitive basis on how humans mentally visualize certain concepts in sentences during language comprehension.

In this paper, we propose an enhanced multi-modal encoder-decoder model, in which the encoder attends to the input sentence and the decoders predict image features and the target sentence. We train the model on images and respective captions from COCO5K dataset \cite{lin2014microsoft}. We augment the state-of-the-art sentence representations with those produced by our model and conduct a series of experiments on transfer tasks to test the quality of sentence representations. Through detailed analysis, we confirm our hypothesis that self-attention help our model produce more feature-rich visually grounded sentence representations. 


\section{Related Work}


\textbf{Sentence Representations.} Since the inception of word embeddings\cite{mikolov2013distributed}, extensive work have emerged for larger semantic units, such as sentences and paragraphs. These works range from deep neural models \cite{kiros2015skip} to log-bilinear models \cite{chen2017efficient, hill2016learning}. A recent work proposed using supervised learning of a specific task as a leverage to obtain general sentence representation \cite{conneau2017supervised}. 

\textbf{Joint Learning of Language and Vision.} Convergence between computer vision and NLP researches have increasingly become common. Image captioning \cite{xu2015show, vinyals2015show, karpathy2015deep, socher2014grounded} and image synthesis \cite{mansimov2015generating} are two common tasks. There have been significant studies focusing on improving word embeddings \cite{lazaridou2015combining, kiela2014learning}, phrase embeddings \cite{krishnamurthy2013jointly}, sentence embeddings \cite{kiela2017learning, chrupala2015learning}, language models \cite{kiros2014unifying} through multi-modal learning of vision and language. Among all studies, \cite{kiela2017learning} is the first to apply skip-gram-like intuition (predicting multiple modalities from langauge) to joint learning of language and vision in the perspective of general sentence representations.

\textbf{Attention Mechanism in Multi-Modal Semantics.} Attention mechanism was first introduced in \cite{bahdanau2014neural} for neural machine translation. Similar intuitions have been applied to various NLP \cite{seo2016bidirectional, chorowski2015attention, lin2017structured} and vision tasks \cite{xu2015show}. \cite{xu2015show} applied attention mechanism to images to bind specific visual features to language. Recently, self-attention mechanism \cite{lin2017structured} has been proposed for situations where there are no extra source of information to ``guide the extraction of sentence embedding''. In this work, we propose a novel sentence encoder for the multi-modal encoder-decoder framework that leverages the self-attention mechanism. To the best of our knowledge, such attempt is the first among studies on joint learning of language and vision.


\section{Proposed Method}

Given a data sample $\left( \mathbf{X}, \mathbf{Y}, \mathbf{h}_I \right) \in \mathcal{D}$, where $\mathbf{X}$ is the source caption, $\mathbf{Y}$ is the target caption, and $\mathbf{h}_I$ is the hidden representation of the image, our goal is to predict $\mathbf{Y}$ and $\mathbf{h}_I$ with $\mathbf{X}$, and the hidden representation in the middle serves as the general sentence representation.

\subsection{Visually Grounded Encoder-Decoder Framework}
\label{sec-ed}

We base our model on the encoder-decoder framework introduced in \cite{kiela2017learning}. A bidirectional Long Short-Term Memory (LSTM) \cite{hochreiter1997long} encodes an input sentence and produces a sentence representation for the input. A pair of LSTM cells encodes the input sequence in both directions and produce two final hidden states: $\vec{\mathbf{h}}_t$ and $\cev{\mathbf{h}}_t$. The hidden representation of the entire sequence is produced by selecting maximum elements between the two hidden states: $\mathbf{h}_S = \max ( \vec{\mathbf{h}}_t, \cev{\mathbf{h}}_t )$. 

The decoder calculates the probability of a target word $\mathbf{y}_t$ at each time step $t$, conditional to the sentence representation $\mathbf{h}_S$ and all target words before $t$. $ P \left( \mathbf{y}_t \mid \mathbf{y}_{<t}, \mathbf{h}_S \right)$.

The objective of the basic encoder-decoder model is thus the negative log-likelihood of the target sentence given all model parameters: $\mathcal{L}_C = - \sum_{\mathbf{X}, \mathbf{Y} \in \mathcal{D}} \sum_{\mathbf{y}_t \in \mathbf{Y}} \log P \left( \mathbf{y}_t \mid \mathbf{y}_{<t}, \mathbf{X}, \Theta \right)$ .

\subsection{Visual Grounding}
\label{sec-vg}

Given the source caption representation $\mathbf{h}_S$ and the relevant image representation $\mathbf{h}_I$, we associate the two representations by projecting $\mathbf{h}_S$ into image feature space. We train the model to rank the similarity between predicted image features $\tilde{\mathbf{h}}_I$ and the target image features $\mathbf{h}_I$ higher than other pairs, which is achieved by ranking loss functions. Although margin ranking loss has been the dominant choice for training cross-modal feature matching \cite{kiros2014unifying, kiela2017learning, lee2017emergent}, we find that \textit{log-exp-sum pairwise ranking} \cite{li2017improving} yields better results in terms of evaluation performance and efficiency. Thus, the objective for ranking 

\begin{equation}
\label{loss_ci}
\mathcal{L}_{VG} = \log \left( 1 + \sum_{\tilde{\mathbf{h}}_I, \mathbf{h}_I} \sum_{\left(\mathbf{h}_I^\prime, \tilde{\mathbf{h}}_I^\prime \right) \in \mathcal{N}}  {\exp \left(sim \left( \tilde{\mathbf{h}}_I^\prime , \mathbf{h}_I^\prime \right) - sim \left( \tilde{\mathbf{h}}_I , \mathbf{h}_I \right) \right) } \right)
\end{equation}


where $\mathcal{N}$ is the set of negative examples and $sim$ is cosine similarity. 

\subsection{Visual Grounding with Self-Attention}


Let $h_t \in \mathbb{R} ^ {d_h}$ be the encoder hidden state at timestep $t$ concatenated from two opposite directional LSTMs ($d_h$ is the dimensionality of sentence representations). Let $H \in \mathbb{R} ^ {d_h \times T}$ be the hidden state matrix where $t$-th column of $H$ is $h_t$. The self-attention mechanism aims to learn attention weight $\alpha_t$, i.e. how much attention must be paid to hidden state $\mathbf{h}_t$, based on all hidden states $H$. Since there could be multiple ways to attend depending on desired features, we allow multiple attention vectors to be learned. Attention matrix $\mathbf{A} \in \mathbb{R} ^ {n_a \times T}$ is a stack of $n_a$ attention vectors, obtained through attention layers: $\mathbf{A} = softmax \left( \mathbf{W}_{a2} \tanh \left( \mathbf{W}_{a1} H \right) \right)$. $\mathbf{W}_{a1} \in \mathbb{R} ^ {d_a \times d_h}$ and $\mathbf{W}_{a2} \in \mathbb{R} ^ {n_a \times d_a}$ are attention parameters and $d_a$ is a hyperparameter. The context matrix $\mathbf{C} \in \mathbb{R} ^ {n_a \times d_h}$ is obtained by $\mathbf{C} = \mathbf{A} \mathbf{H}$. Finally, we compress the context matrix into a fixed size representation $\mathbf{h}_A$ by max-pooling all context vectors: $
\mathbf{h}_A = \max \left(\mathbf{c}_1, \mathbf{c}_2, \ldots, \mathbf{c}_{n_a} \right) $. Attended representation $\mathbf{h}_A$ and encoder-decoder representation $\mathbf{h}_S$ are concatenated into the final self-attentive sentence representation $\mathbf{h}$. This hybrid representation replaces $\mathbf{h}_S$ and is used to predict image features (Section \ref{sec-vg}) and target caption (Section \ref{sec-ed}).

\subsection{Learning Objectives}

Following the experimental design of \cite{kiela2017learning}, we conduct experiments on three different learning objectives: \textsc{Cap2All}, \textsc{Cap2Cap}, \textsc{Cap2Img}. Under \textsc{Cap2All}, the model is trained to predict both the target caption and the associated image: $\mathcal{L} = \mathcal{L}_{C} + \mathcal{L}_{VG}$. Under \textsc{Cap2Cap}, the model is trained to predict only the target caption ($\mathcal{L} = \mathcal{L}_C$) and, under \textsc{Cap2Img}, only the associated image ($\mathcal{L} = \mathcal{L}_{VG}$). 
\section{Experiments} 

\subsection{Implementation Details}
\label{sec-imp}

Word embeddings $\mathbf{W}_E$ are initialized with GloVe \cite{pennington2014glove}. The hidden dimension of each encoder and decoder LSTM cell ($d_h$) is 1024\footnote{However, for baseline models (without self-attention), we use $d_h = 2048$ to match the dimensionality (2048) of sentence representations produced by our proposed models.}. We use Adam optimizer \cite{kingma2014adam} and clip the gradients to between -5 and 5. Number of layers, dropout, and non-linearity for image feature prediction layers are 4, 0.3 and ReLU \cite{nair2010rectified} respectively. Dimensionality of hidden attention layers ($d_a$) is 350 and number of attentions ($n_a$) is 30. We employ orthogonal initialization \cite{saxe2014exact} for recurrent weights and xavier initialization \cite{glorot2010understanding} for all others. For the datasets, we use Karpathy and Fei-Fei's split for MS-COCO dataset \cite{karpathy2015deep}. Image features are prepared by extracting hidden representations at the final layer of ResNet-101 \cite{he2016deep}. We evaluate sentence representation quality using \verb|SentEval|\footnote{https://github.com/facebookresearch/SentEval} \cite{conneau2017supervised, kiela2017learning} scripts. Mini-batch size is 128 and negative samples are prepared from remaining data samples in the same mini-batch.

\subsection{Evaluation}

Adhering to the experimental settings of \cite{kiela2017learning}, we concatenate sentence representations produced from our model with those obtained from the state-of-the-art unsupervised learning model (Layer Normalized Skip-Thoughts, ST-LN) \cite{ba2016layer}. We evaluate the quality of sentence representations produced from different variants of our encoders on well-known transfer tasks: movie review sentiment (MR) \cite{pang2005seeing}, customer reviews (CR) \cite{hu2004mining}, subjectivity (SUBJ) \cite{pang2004sentimental}, opinion polarity (MPQA) \cite{wiebe2005annotating}, paraphrase identification (MSRP) \cite{dolan2004unsupervised}, binary sentiment classification (SST) \cite{socher2013recursive}, SICK entailment and SICK relatedness \cite{marelli2014semeval}. 

\subsection{Results}

Results are shown in Table \ref{clf-results}. Results show that incorporating self-attention mechanism in the encoder is beneficial for most tasks. However, original models were better in some tasks (CR, MPQA, MRPC), suggesting that self-attention mechanism could sometimes introduce noise in sentence features. Overall, utilizing self-attentive sentence representation further improves performances in 5 out of 8 tasks. Considering that models with self-attention employ smaller LSTM cells (1024) than those without (2048) (Section \ref{sec-imp}), the performance improvements are significant. Results on COCO5K image and caption retrieval tasks (not included in the paper due to limited space) show comparable performances to other more specialized methods \cite{karpathy2015deep, klein2015associating}.

\begin{table}[htb]
	\caption{Classification performance on transfer tasks. We report F1-score for MRPC, Pearson coefficient for SICK-R and accuracy for most others. All sentence representations have been concatenated with ST-LN embeddings. Note that the discrepancy between results reported in this paper and the referenced paper is likely due to differences in minor implementation details and experimental environment. Our models are denoted by $\dagger$.}
	\label{clf-results}
	\centering
	\begin{tabular}{llllllllll}
		\toprule
		Method     					& MR		& CR		& SUBJ		& MPQA		& MRPC		& SST & SICK-E & SICK-R	\\
		\midrule
		ST-LN  \cite{ba2016layer}    & 75.46  	& 76.98		& 92.60		& 86.46		& 82.23		& 82.26 & 80.76 & 84.39 \\
		\midrule
		\textsc{Cap2Cap}  \cite{kiela2017learning} & 75.45 &	\textbf{77.85} &	92.84 &	\textbf{87.45} &	81.92 &	82.54 & 80.98 & 83.62\\
		
		\textsc{Cap2Img}  \cite{kiela2017learning} & 75.81	& 77.35 &	92.59 &	86.99 &	73.16 &	82.43 & 81.25 & 81.59\\
		
		\textsc{Cap2All}  \cite{kiela2017learning} & 75.92 &	77.46 &	92.86 &	87.04 &	\textbf{82.26} &	81.16 & 81.59 & 84.37\\
		\midrule
		Att. \textsc{Cap2Img} $^\dagger$ & 75.88	& 77.16	& \textbf{92.91} & 86.57	& 82.16 & \textbf{83.03} & \textbf{81.69} & 83.95 \\
		Att. \textsc{Cap2All} $^\dagger$ & \textbf{75.98} &	77.43 &	92.44 &	86.33 &	81.88 &	81.60 & 81.08 & \textbf{84.54} \\
		\bottomrule
	\end{tabular}
\end{table}

\subsection{Attention Mechanism at Work}

\begin{figure}[htb]
	\centering
	\includegraphics[width=1.0\textwidth,height=3cm]{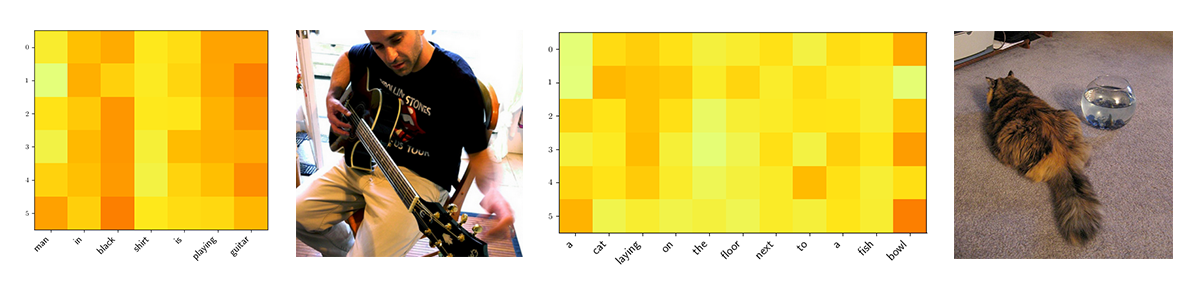}
	\caption{Activated attention weights on two samples from MS-COCO dataset. Vertical axis shows attention vectors learned by our model (compressed due to space limit). Note how the sentence encoder learned to identify words with strong visual associations.}
	\label{fig:attention}
\end{figure}

In order to study the effects of incorporating self-attention mechanism in joint prediction of image and language features, we examine attention vectors for selected samples from MS-COCO dataset and compare them to associated images (Figure \ref{fig:attention}). For example, given the sentence ``man in black shirt is playing guitar'', our model identifies words that have association with strong visual imagery, such as ``man'', ``black'' and ``guitar''. Given the second sentence, our model learned to attend to visually significant words such as ``cat'' and ``bowl''. These findings show that visually grounding self-attended sentence representations helps to expose word-level visual features onto sentence representations \cite{kiela2017learning}.  

\section{Conclusion and Future Work}

In this paper, we proposed a novel encoder that exploits self-attention mechanism. We trained the model using MS-COCO dataset and evaluated sentence representations produced by our model (combined with universal sentence representations) on several transfer tasks. Results show that the self-attention mechanism not only improves the qualities of general sentence representations but also guides the encoder to emphasize certain visually associable words, which helps to make visual features more prominent in the sentence representations. As future work, we intend to explore cross-modal attention mechanism to further intertwine language and visual information for the purpose of improving sentence representation quality. 

\subsubsection*{Acknowledgments}

This work was supported by BK21 Plus for Pioneers in Innovative Computing(Dept. of Computer Science and Engineering, SNU) funded by National Research Foundation of Korea(NRF) (21A20151113068). Also, this work would not be possible without invaluable discussions with knowledgeable and helpful colleagues. 

\nocite{*}
\bibliographystyle{ieeetr}
\bibliography{ref}

\end{document}